\documentclass{article}
\usepackage{xcolor}
\usepackage{subcaption}
\usepackage{amssymb}
\usepackage{booktabs}
\usepackage{float}
\usepackage{graphicx}
\usepackage{amsmath}
\usepackage{multirow}
\usepackage{wrapfig}
\usepackage{threeparttable}
\usepackage{tabularx}
\usepackage{longtable}
\usepackage{array}
\usepackage{placeins}
\usepackage{algpseudocode}
\usepackage[table]{xcolor}

\usepackage{hyperref}
\usepackage[final]{corl_2026} 

\title{\methodname{}: Unified Multi-Skill Policy Learning for Humanoid Soccer}

\definecolor{modAcolor}{RGB}{245, 60, 68}
\definecolor{modBcolor}{RGB}{36,144,135}
\definecolor{modCcolor}{RGB}{72,116,233}

\newcommand{\modA}[1]{\textcolor{modAcolor}{#1}}
\newcommand{\modB}[1]{\textcolor{modBcolor}{#1}}
\newcommand{\modC}[1]{\textcolor{modCcolor}{#1}}

\author{
\begin{tabular}{cccc}
\textbf{Zhangchen Ye\textsuperscript{1, 2*}} &
\textbf{Enxuan Ruan\textsuperscript{1*}} &
\textbf{Yifei Bao\textsuperscript{1*}} &
\textbf{Runhan Huang\textsuperscript{2}}
\end{tabular}
\\[0.2em]
\begin{tabular}{ccccc}
\textbf{Jiankun Yang\textsuperscript{1}} &
\textbf{Jiakang Jin\textsuperscript{1}} &
\textbf{Yixiao Huo\textsuperscript{1}} &
\textbf{Pengyuan Wang\textsuperscript{1}} &
\textbf{Yinan Han\textsuperscript{1}}
\end{tabular}
\\[0.2em]
\begin{tabular}{cccc}
\textbf{Huaxing Huang\textsuperscript{1}} &
\textbf{Wenhao Cui\textsuperscript{1}} &
\textbf{Yiming Li\textsuperscript{2\dag}} &
\textbf{Xiaoyu Tian\textsuperscript{1\dag}}
\end{tabular}
\\[0.8em]
\textsuperscript{1} Noetix Robotics;
\textsuperscript{2} Tsinghua University
\\
* Equal Contribution
\quad
\dag~Corresponding Author
}

\newcommand{\methodname}{SkillX}

\begin{document}
\maketitle

\begin{center}
    \vspace{-2em}
    Project page: \href{https://yzc0731.github.io/SkillX/}{\texttt{yzc0731.github.io/SkillX}}
\end{center}

\begin{figure}[htbp]
\vspace{-1em}
    \centering
    \includegraphics[width=\textwidth]{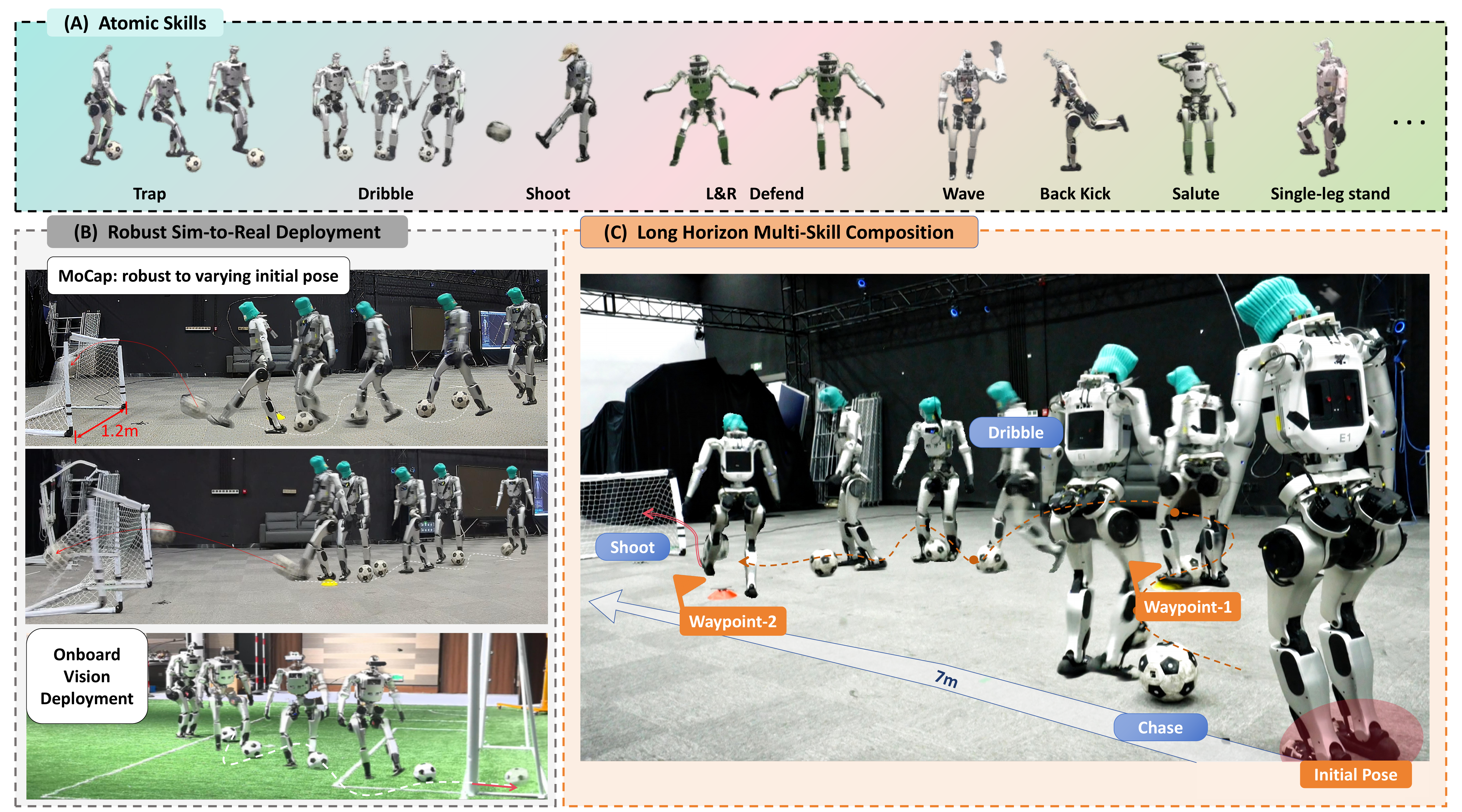}
    \caption{Our framework \textbf{\methodname{}} enables a humanoid robot to execute diverse soccer skills and compose them into long-horizon behaviors.
    \textbf{(A)} The robot learns \textit{atomic skills} such as trapping, dribbling, shooting, and additional soccer skills.
    \textbf{(B)} The learned policy achieves robust \textit{sim-to-real deployment} on the Noetix E1 robot using both the motion-capture system and onboard vision.
    \textbf{(C)} The robot composes atomic skills and performs \textit{long-horizon multi-skill tasks}.
    }
    \label{fig:teaser}
\end{figure}
\begin{abstract}

    Humanoid soccer is a challenging testbed for dynamic whole-body control, requiring robots to coordinate balance, locomotion, object interaction, and skill switching over long horizons. Existing humanoid sports methods often rely on task-specific multi-stage pipelines, making it difficult to jointly learn and compose multiple object-interactive skills within a single deployable policy. To address this, we present SkillX, a unified reinforcement learning framework that learns and composes multiple atomic soccer skills through a single command-conditioned policy. SkillX integrates three core designs: skill-specific adversarial motion priors, skill-specific critics, and an object-aware temporal encoder, enabling the robot to execute atomic skills and transition among them such as dribbling, trapping, and shooting. Experiments in simulation and on a real Noetix E1 humanoid demonstrate robust multi-skill execution, long-horizon skill composition, and successful sim-to-real deployment.

\end{abstract}



\section{Introduction}

Humanoid soccer~\cite{kitano1997robocup, gerndt2015humanoid, yi2016hierarchical, da2021deep, tirumala2024soccervision, haarnoja2024learning} is a challenging testbed for dynamic whole-body control because it requires a robot to coordinate balance, locomotion, object interaction, and skill switching. 
A soccer robot must execute athletic skills such as dribbling, shooting, and trapping, while composing them into long-horizon behaviors in response to a moving ball and changing task objectives, going beyond conventional single-skill motion tracking, in which a policy primarily follows reference motions. 

Recent progress in humanoid control has been largely driven by motion tracking, which enables agile locomotion and stylized motion imitation~\cite{ze2025twist, xie2026kungfubot, liao2025beyondmimic, luo2025sonic, ma2026robust}. This paradigm has also been extended to athletic ball interaction tasks such as soccer~\cite{kong2026learning}, basketball~\cite{wang2026humanx}, tennis~\cite{zhang2026learning}, table tennis~\cite{ren2026smash}, and badminton~\cite{liu2025humanoid, chen2026learning}. 
However, existing methods often rely on task-specific multi-stage pipelines that combine motion imitation with task-oriented refinement, making it difficult to jointly learn and compose multiple object-interactive skills with a single deployable policy.

As a representative instance of multi-skill object-interactive control, unified humanoid soccer further amplifies these difficulties through three coupled challenges.
First, different skills exhibit \textit{motion-style heterogeneity}, involving distinct kinematic patterns and contact behaviors. Naive joint training can cause cross-skill interference and unreliable transitions~\cite{huang2025moe, huang2025learning, ma2026cmoe}. Second, these skills involve \textit{skill-dependent task objectives}, making shared value estimation difficult across skill modes. Third, soccer interactions are subject to \textit{partial observability}: ball velocity, contact events, and short-term interaction outcomes are noisy or latent and must be inferred from observation histories. Together, these challenges make it difficult for a single policy to preserve skill-specific motion quality while composing skills robustly over long horizons.

We present \textbf{\methodname{}}, a unified reinforcement learning framework for multi-skill humanoid soccer. 
As illustrated in Fig.~\ref{fig:teaser}, the learned policy supports diverse atomic skills, robust sim-to-real deployment, and long-horizon multi-skill composition.
First, \textit{skill-specific adversarial motion priors} separate motion regularization across skills to reduce style interference. 
Second, \textit{skill-specific critics} address skill-dependent value estimation across
different task objectives. Third, a transformer-based \textit{object-aware temporal encoder} addresses partial observability by estimating hidden robot and object states from observation histories. 
Together with adaptive command sampling, a command-duration curriculum, and object domain randomization, these components enable stable command-driven skill composition with a single deployed policy. Although instantiated in humanoid soccer, these designs target general challenges in multi-skill object-interactive control.

We evaluate \methodname{} on humanoid soccer tasks in simulation and on the real Noetix E1 humanoid robot.
In simulation, we test both atomic skills and composite tasks of varying difficulty levels, compare against representative baselines, and conduct ablations to quantify the contribution of each major component.
On the hard long-horizon trap-dribble-shoot task, \methodname{} achieves \textbf{81.7\%} success, compared with 27.4\% for AMP. 
On hardware, we deploy the learned policy with a motion-capture (MoCap) system and further study onboard-vision deployment using a head-mounted camera.
The real robot achieves a \textbf{72.5\%} average success rate across four soccer tasks, demonstrating reliable object-interaction skill execution and
command-driven skill transitions.

Our contributions are summarized as follows:
\begin{itemize}
    \item We propose \textbf{\methodname{}}, a unified reinforcement learning framework for \textbf{multi-skill humanoid soccer}, enabling a single policy to execute multiple atomic skills and compose them into long-horizon tasks.
    \item We introduce multi-skill training mechanisms, including \textbf{skill-specific adversarial motion priors}, \textbf{skill-specific critics}, and a \textbf{transformer-based object-aware temporal encoder}, to address motion-style heterogeneity, skill-dependent task objectives, and partial observability in multi-skill object-interactive control.
    \item We validate \methodname{} in both simulation and real-world deployment, demonstrating robust sim-to-real deployment and \textbf{long-horizon multi-skill composition}. 
\end{itemize}

\section{Related Work}
\subsection{Multi-Task Reinforcement Learning in Locomotion}
Multi-task reinforcement learning methods for locomotion aim to train a unified policy for diverse motion skills, but often face gradient interference across heterogeneous tasks~\cite{huang2025moe,wang2025more,ma2026cmoe}. Major solution categories include hierarchical policies, implicit skill disentanglement, and mixture-of-experts (MoE) architectures. Hierarchical methods pre-train individual skills and use high-level modules for skill selection and transitions~\cite{shah2023mtac,kuang2025skillblender,Wang_2025_CVPRSkillMimic,yu2026discovery}, but typically require multi-stage training and do not readily support end-to-end optimization. Implicit disentanglement methods regularize diverse motion styles in a shared latent space~\cite{peng2022ase,tessler2023calm,huang2025learning}, but can still suffer from latent entanglement, making it difficult to fully separate skill-specific features. MoE methods use gating mechanisms to mitigate gradient conflicts in end-to-end multi-skill learning~\cite{huang2025moe, ma2026cmoe}, but can suffer from expert imbalance or unstable specialization. Inspired by multi-adversarial motion priors~\cite{vollenweider2023advanced} and multi-critic reinforcement learning~\cite{mysore2022multi}, our method decouples style regularization and value estimation across skills during training while maintaining a single unified policy for deployment, enabling command-driven skill execution and smooth transitions.

\subsection{Humanoid Athletic Ball Interaction}
While multi-skill locomotion studies focus on learning diverse motions, athletic ball interaction introduces an additional challenge of coordinating balance and whole-body motion under fast-changing object dynamics and contact-rich interactions. Recent learning-based methods have made considerable progress in humanoid ball interaction, yet many existing studies primarily target isolated skills, such as flat strikes in table tennis~\cite{su2025hitter}, soccer dribbling~\cite{wang2025dribble}, and shooting~\cite{wang2025learning,xu2025learning,kong2026learning}. To equip robots with broader athletic capabilities, prior methods often rely on multi-stage training pipelines, including progressive skill refinement~\cite{liu2025humanoid,chen2026learning} and policy distillation~\cite{wang2026humanx,zhang2026learning}. While effective, these designs can require careful training schedules and additional optimization procedures. 
In contrast, we study humanoid soccer as a multi-skill ball-interaction setting, in which the policy must compose multiple ball-control skills while inferring interaction-relevant object states from partial observations. Our framework addresses these requirements by integrating skill-specialized training signals and an object-aware temporal encoder into a unified policy.

\section{Method}
\begin{figure*}[t]
    \centering
    \vspace{2mm}
    \includegraphics[width=\textwidth]{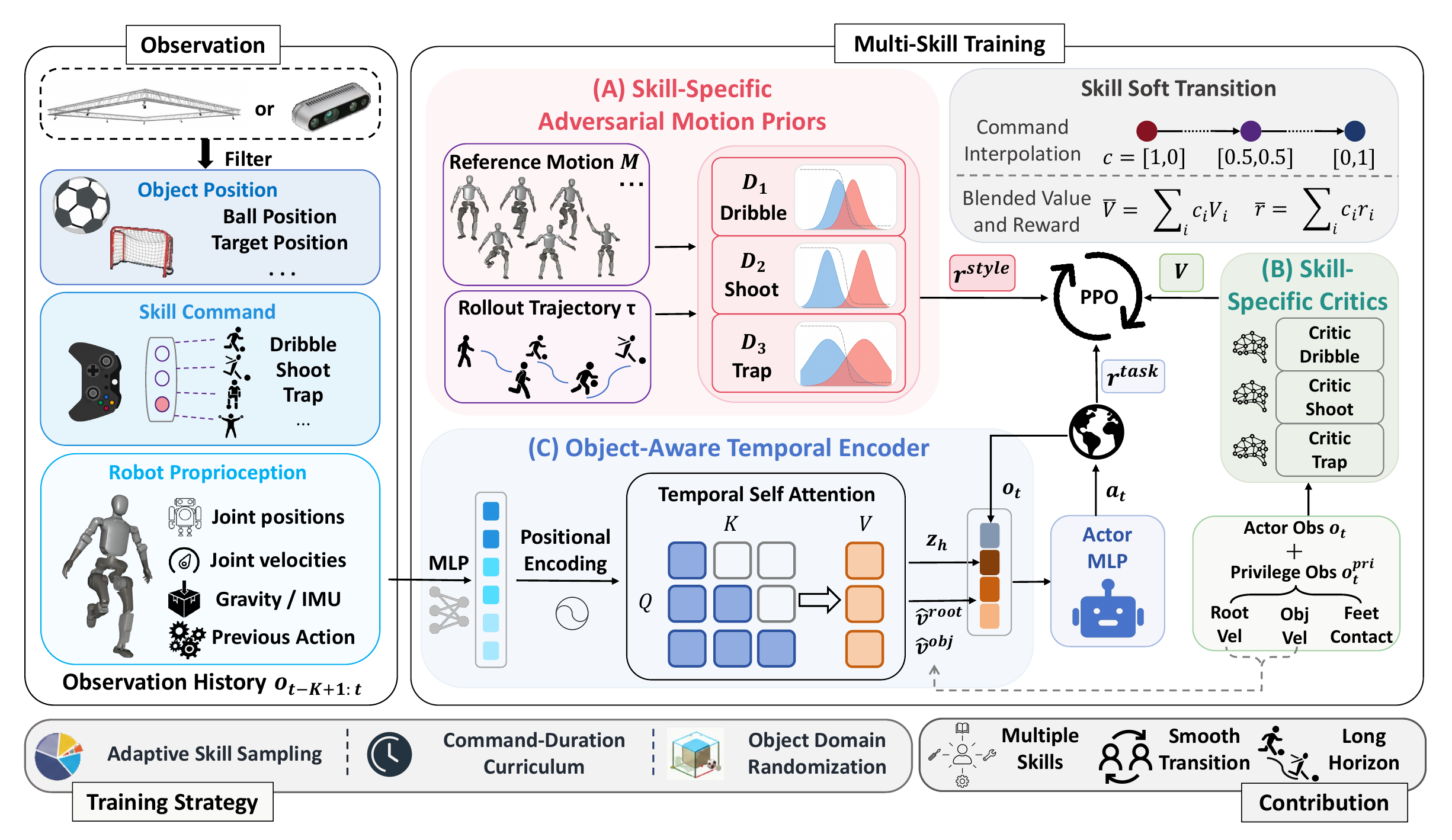}
    \caption{\textbf{Overview of \methodname{}.}
The policy receives a history of robot proprioception, object states, and skill
commands, and outputs joint-position targets. 
\textbf{(A)} \modA{\textit{Skill-Specific Adversarial Motion Priors}} compare policy rollouts with reference motions and provide style rewards for heterogeneous skills.
\textbf{(B)} \modB{\textit{Skill-Specific Critics}} estimate skill-dependent values; during
soft skill transitions, the command vector interpolates both rewards and values
for PPO updates.
\textbf{(C)} \modC{\textit{Object-Aware Temporal Encoder}} aggregates observation histories
with temporal self-attention and provides compact history features for the actor.
}
    \label{fig:pipeline}
\end{figure*}

We design \methodname{} as a unified multi-skill reinforcement learning
framework, instantiated here for humanoid soccer. The key principle is to keep deployment simple with a single command-conditioned actor, while using skill-specific training signals to reduce interference across heterogeneous skills. As shown in Fig.~\ref{fig:pipeline}, the framework consists of three core components: \textbf{(A)} skill-specific adversarial motion priors for preserving distinct motion styles, \textbf{(B)} skill-specific critics for skill-dependent value estimation, and \textbf{(C)} an object-aware temporal encoder for estimating hidden robot and object states required for precise interaction. We detail these components in Sec.~\ref{sec:amp}, Sec.~\ref{sec:critic}, and Sec.~\ref{sec:him}, respectively.

\subsection{Problem Formulation}
We formulate multi-skill humanoid soccer as an MDP
\(\mathcal{M}=(\mathcal{S},\mathcal{A},P,r,\gamma)\) and train a policy
\(\pi_\theta\) with PPO~\cite{schulman2017proximal} to maximize the expected discounted return. Since the actor receives only partial proprioceptive and object observations, the policy is conditioned on a history window of length \(K\). The instantaneous observation at time \(t\) is
$\mathbf{o}_t =
    \left[
    \mathbf{p}_{t},
    \mathbf{c}_{t},
    \mathbf{o}^{\mathrm{ext}}_{t}
    \right]$, 
where \(\mathbf{p}\) denotes proprioception, \(\mathbf{c}\) the skill command,
and \(\mathbf{o}^{\mathrm{ext}}\) external object information. In our soccer
task, \(\mathbf{o}^{\mathrm{ext}}\) contains the robot-centric relative
positions of the ball and goal. The command vector
\(\mathbf{c}_t\in[0,1]^{N_{\mathrm{skill}}}\) satisfies
\(\sum_i c_{t,i}=1\): it is one-hot during single-skill execution and linearly
interpolated during skill switches to produce soft transitions. The policy
outputs target joint positions executed by low-level PD controllers. Detailed
observation specifications are provided in Appendix~\ref{app:obs_detail}.

\subsection{Skill-Specific Adversarial Motion Priors}
\label{sec:amp}
Different skills in a unified policy can exhibit distinct motion distributions
and contact patterns.
While AMP~\cite{peng2021amp} uses an adversarial discriminator to provide motion-style rewards, a shared discriminator in a multi-skill setting can produce ambiguous style guidance and weaken skill-specific style separation. We address this issue with a set of skill-specific adversarial motion priors $\{D_i\}_{i=1}^{N_{\mathrm{skill}}}$, where each discriminator is trained for one atomic skill while the actor remains shared across all skills. 

Each discriminator $D_i$ distinguishes expert motion segments of skill $i$ from policy-generated segments under the same command mode.
Its input is a compact observation consisting of motion-centric features, including joint states, key body positions, root velocities, and foot contact states. 
Following AMP~\cite{peng2021amp}, we train each $D_i$ with a least-squares GAN objective and gradient penalty:
\begin{equation}
\mathcal{L}_{D_i}
=
\mathbb{E}_{\boldsymbol{\tau}\sim d_i^M}
\left[(D_i(\boldsymbol{\tau})-1)^2\right]
+
\mathbb{E}_{\boldsymbol{\tau}\sim d_i^\pi}
\left[(D_i(\boldsymbol{\tau})+1)^2\right]
+
\frac{w^{\mathrm{gp}}}{2}
\mathbb{E}_{\boldsymbol{\tau}\sim d_i^M}
\left\|\nabla_{\boldsymbol{\tau}} D_i(\boldsymbol{\tau})\right\|_2^2,
\end{equation}
where $\boldsymbol{\tau}$ denotes a multi-step motion segment, \(d^M_i\) is
the expert motion distribution for skill \(i\), and \(d^\pi_i\) is the
corresponding policy-generated distribution.

The style reward produced by discriminator \(D_i\) is defined as
\begin{equation}
r_i^{\mathrm{style}}(t)
=
\max\left[
0,\;
1 - 0.25\left(D_i(\boldsymbol{\tau}^{\pi}_t)-1\right)^2
\right],
\end{equation}
where \(\boldsymbol{\tau}^{\pi}_t\) is the policy motion segment ending at time
\(t\). For one-hot command segments, the policy receives the style reward from the active skill discriminator. 
During a soft transition, we compute a
blended style reward according to the interpolated command vector $\mathbf{c}_{t}$:
\begin{equation}
    \bar{r}^{\mathrm{style}}(t)
    =
    \sum_{i=1}^{N_{\mathrm{skill}}}
    c_{t,i}\,r_i^{\mathrm{style}}(t).
\end{equation}
We refer to this command-weighted interpolation as \textit{command blending}.

Motion segments spanning command switches are excluded from the discriminator
updates, since they contain mixed behaviors and should not be treated as clean
samples of any single skill.

\subsection{Multi-Critic PPO with Skill-Specific Critics}
\label{sec:critic}

In standard PPO, a single critic estimates one value function for all training samples. 
This is insufficient for command-driven multi-skill control, where the same physical state can have different expected returns under different skill objectives. 
Fitting these skill-dependent returns with one shared critic can introduce value interference and biased advantage estimates.
To address this issue, we maintain a set of skill-specific critics \(\{V_i\}_{i=1}^{N_{\mathrm{skill}}}\), while keeping a single shared actor. 

Given the command vector \(\mathbf{c}_t\), we compute a command-conditioned value estimate by selecting or blending the skill-specific critics:
\begin{equation}
    \bar{V}(t)
    =
    \sum_{i=1}^{N_{\mathrm{skill}}}
    c_{t,i} V_i(t).
\end{equation}
For one-hot commands, \(\bar{V}(t)\) selects the active skill critic; during soft transitions, it uses the same command blending rule used for the task and style rewards. PPO advantages are then computed using the command-conditioned reward and value estimates, followed by the standard PPO update.

In practice, the critics share a common feature extractor and use separate skill-specific value heads. This parameter-efficient design allows each head to specialize in the value function of one skill, rather than forcing a single critic to represent all skill-dependent returns. As a result, the critics provide more accurate command-conditioned value estimates, leading to better advantage estimation and more stable multi-skill policy optimization.

\subsection{Object-Aware Temporal Encoder}
\label{sec:him}
Dynamic object-interaction tasks require the policy to react to fast-changing
robot-object states, while receiving partial and noisy
observations. 
We therefore introduce an object-aware temporal encoder that aggregates observation histories and produces compact features for precise interaction. 

Given the observation history \(\mathbf{o}_{t-K+1:t}\), we first embed each
observation with an MLP and add sinusoidal positional encoding. The resulting
sequence is processed by a causal transformer with temporal self-attention:
\begin{align}
    \mathbf{E}_{t-K+1:t}
    &=
    \phi_{\mathrm{emb}}(\mathbf{o}_{t-K+1:t}) + \mathbf{PE}, \\
    \mathbf{H}_{t-K+1:t}
    &=
    \mathcal{T}_{\phi}(\mathbf{E}_{t-K+1:t}, \mathbf{M}_K),
\end{align}
where \(\mathbf{PE}\) is the positional encoding and \(\mathbf{M}_K\) is a
causal attention mask that prevents future information leakage. The history latent \(\mathbf{z}_h\) is obtained by applying layer normalization and average pooling over the temporal dimension of \(\mathbf{H}_{t-K+1:t}\).

Following the hybrid optimization idea of HIM~\cite{long2024hybrid}, we train the temporal encoder with two auxiliary objectives. First, we use a
self-supervised temporal consistency loss based on Barlow Twins to regularize the history latent under temporally shifted observation windows. Second, we attach a lightweight estimation head to predict hidden states that are important
for interaction:
\begin{equation}
    \hat{\mathbf{v}}
    =
    h_{\mathrm{est}}(\mathbf{z}_h),
    \qquad
    \hat{\mathbf{v}}
    =
    \left[
    \hat{\mathbf{v}}^{\mathrm{root}},
    \hat{\mathbf{v}}^{\mathrm{obj}}
    \right].
\end{equation}
Unlike purely proprioceptive internal models, the additional object-velocity target makes the latent representation sensitive to ball motion, which is available as privileged supervision in simulation but must be inferred from noisy position histories during deployment. The final encoder objective is \(\mathcal{L}_{\mathrm{enc}}=\mathcal{L}_{\mathrm{est}}+\mathcal{L}_{\mathrm{bt}}\), where \(\mathcal{L}_{\mathrm{est}}\) penalizes root- and object-velocity prediction errors and \(\mathcal{L}_{\mathrm{bt}}\) denotes the Barlow-Twins loss. The resulting latent \(\mathbf{z}_h\) and estimated velocities \(\hat{\mathbf{v}}\) are provided to the actor as compact history-based features for precise robot-object interaction under noisy or delayed observations.

\subsection{Training Strategy}
To further improve stability, sample efficiency, and generalization for multi‑skill training, we incorporate several practical training strategies. \textbf{Adaptive command sampling} allocates more rollouts to skills with higher recent failure rates. We track each skill's failure rate using an exponential moving average and update the skill-level sampling probability. The \textbf{command duration curriculum} starts with long single-skill segments and progressively shortens command durations to encourage frequent skill switching.
\textbf{Object domain randomization} perturbs ball mass, friction, and
restitution to improve robustness to interaction variations. Details are
provided in Appendix~\ref{app:strategy}.

\newcommand{\best}[1]{\textbf{#1}}
\newcommand{\na}{\textemdash}

\begin{table*}[t]
\centering
\scriptsize

\label{tab:sequential_skill_eval}
\setlength{\tabcolsep}{5.5pt}
\setlength{\aboverulesep}{0pt}
\setlength{\belowrulesep}{0pt}

\newcommand{\icondribble}{\includegraphics[height=1.4ex]{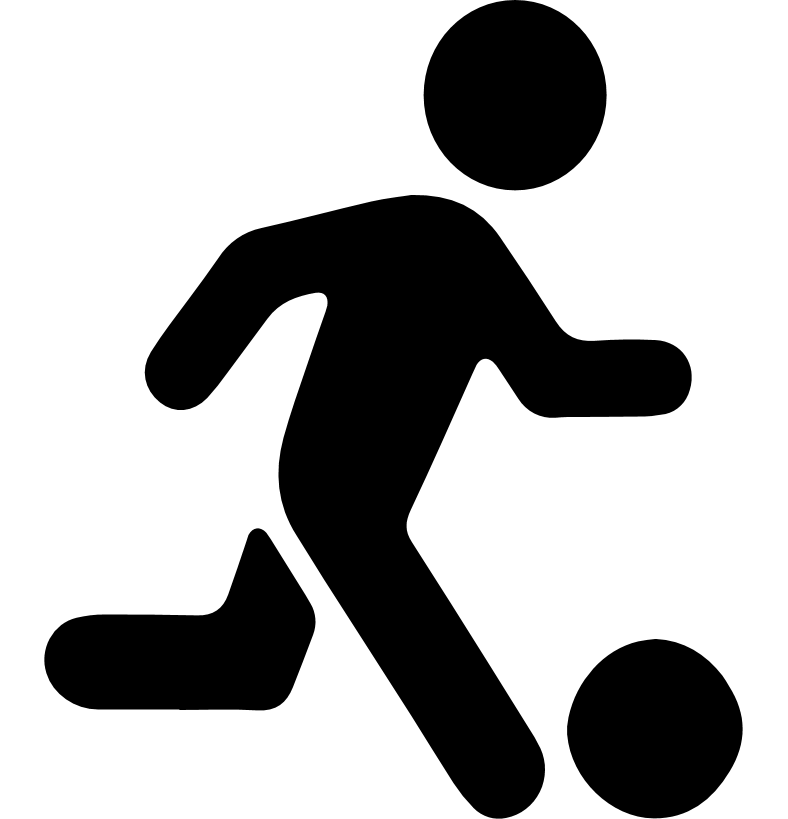}\,}
\newcommand{\iconshoot}{\includegraphics[height=1.4ex]{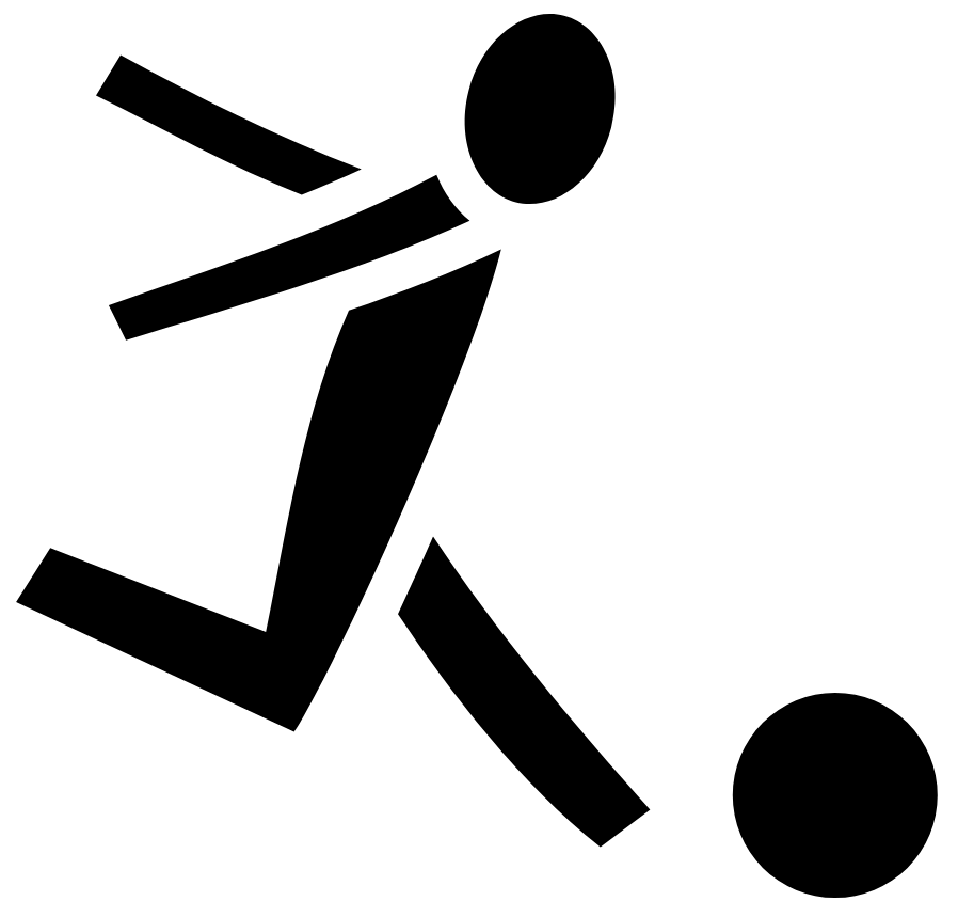}\,}
\newcommand{\icontrap}{\includegraphics[height=1.4ex]{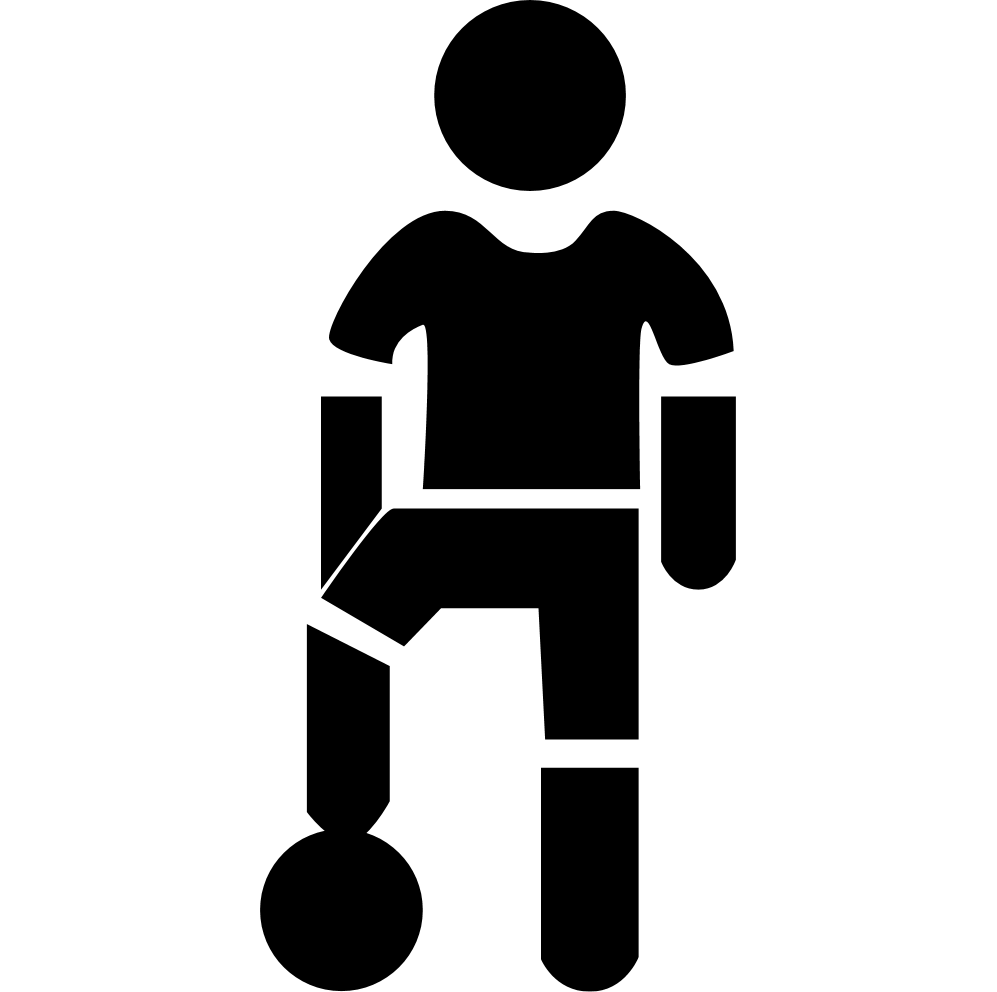}\,}

\begin{threeparttable}
\begin{tabular}{l cccc ccccc}
\toprule
\multirow{2}{*}{Method}
& \multicolumn{4}{c}{Medium}
& \multicolumn{5}{c}{Hard} \\
\cmidrule(lr){2-5}
\cmidrule(lr){6-10}
& \icondribble Dribble
& \iconshoot Shoot
& Overall$\uparrow$
& Time$\downarrow$
& \icontrap Trap
& \icondribble Dribble
& \iconshoot Shoot
& Overall$\uparrow$
& Time$\downarrow$ \\
\midrule
Vanilla PPO & 0.0 & \na & 0.0 & \na & 99.8 & 0.0 & \na & 0.0 & \na \\
AMP & 89.0 & 57.3 & 51.0 & 12.6 & 100.0 & 52.0 & 52.6 & 27.4 & 30.4\\
Conditional AMP & 13.4 & 83.6 & 11.2 & 17.6 & 100.0 & 0.0 & \na & 0.0 & \na \\
MoE-based AMP & 58.0 & 73.4 & 42.6 & 16.6 & 100.0 & 5.0 & 100.0 & 5.0 & 32.6 \\
MoE-Encoder AMP & 100.0 & 67.2 & 67.2 & 13.6 & 100.0 & 95.7 & 47.6 & 45.6 & 34.7 \\
\rowcolor{gray!25}
\textbf{Ours} & 100.0 & 88.0 & \best{88.0} & \best{12.0} & 100.0 & 100.0 & 81.7 & \best{81.7} & \best{28.4} \\
\bottomrule
\end{tabular}
\end{threeparttable}

\caption{
\textbf{Composite Task Evaluation in Simulation.}
Medium: \textsc{Dribble}$\rightarrow$\textsc{Shoot}.
Hard: \textsc{Trap}$\rightarrow$\textsc{Dribble}$\times 3\rightarrow$\textsc{Shoot}.
Atomic skill evaluation is reported in Appendix~\ref{app:sim_results}. Substage success rates are conditional on reaching the corresponding substage; ``Overall'' denotes full-task success. ``\na'' indicates that no episode reaches the corresponding substage.
}
\label{tab:sim_results}
\end{table*}

\section{Experiments}
\label{sec:result}
\subsection{Implementation Details}

We train all policies in NVIDIA Isaac Sim and Isaac Lab~\cite{mittal2025isaaclab}, enabling high-throughput parallel simulation. 
The policy runs at a control frequency of 50 Hz, while the simulation runs at 200 Hz. 
Human reference soccer motions are collected via a MoCap system and retargeted to the 25-DoF Noetix E1 humanoid using PHC~\cite{Luo2023PerpetualHC}.
Real‑world experiments are conducted on the 25‑DoF Noetix E1 humanoid robot. We evaluate two perception pipelines to obtain ball‑related states.
(1) A motion‑capture system provides a 6‑DoF base pose of the humanoid robot and the 3D positions of the ball and goal.
(2) An onboard-vision setup uses a head-mounted ZED2i camera with visual-inertial odometry and YOLOv8-based ball detection~\cite{varghese2024yolov8}. Additional details are provided in Appendix~\ref{app:deploy}.

\subsection{Simulation Results}
\textbf{Evaluation Protocol.} We evaluate policies on both atomic soccer skills and skill-composition tasks. Atomic tasks include isolated shooting, dribbling, and trapping, with results reported in Appendix~\ref{app:sim_results}. In the main text, we focus on two composite tasks with increasing difficulty. The \textit{Medium} task requires \textsc{Dribble}\(\rightarrow\)\textsc{Shoot}, while the \textit{Hard} task requires \textsc{Trap}\(\rightarrow\)\textsc{Dribble}\(\times 3\rightarrow\)\textsc{Shoot}.
In the composite tasks, skill commands are issued by a predefined event-based scheduler: command transitions are triggered when the current subtask objective is completed.
All tasks use randomized initial conditions. The detailed environment parameters, switching conditions, success criteria, and termination conditions are provided in Appendix~\ref{app:task_detail}.

\textbf{Metrics.}
For composite tasks, we report the overall success rate and average completion time, where time is averaged over successful episodes. We also report substage success rates, which are computed over episodes that reach the corresponding substage. All composite-task results in simulation are evaluated over 1000 trials per method.

\textbf{Baseline Setting.} We select representative single-stage training methods as comparative baselines. (1) \textbf{AMP}~\cite{peng2021amp} uses a single shared
adversarial discriminator for all reference motions. (2) \textbf{Conditional AMP}~\cite{huang2025learning, tessler2023calm} is a latent-based multi-skill method. It maps the skill command to a target latent, while a skill discriminator predicts the latent from motion trajectories and provides a cosine-similarity skill reward. (3) \textbf{MoE-based AMP}~\cite{huang2025moe} uses the same shared AMP discriminator as standard AMP, but replaces the shared policy backbone with a mixture-of-experts architecture, where the number of experts matches the number of skill categories. (4) \textbf{MoE-Encoder AMP}~\cite{wu2026toward} augments AMP with a mixture-of-experts history encoder using soft-gated dense routing. (5) \textbf{Vanilla PPO}~\cite{schulman2017proximal} is trained only with task rewards, without adversarial motion priors. For a fair comparison, all methods use the same task rewards, maximum training budget, and evaluation protocol. Additional implementation details and hyperparameters are provided in Appendix~\ref{app:baseline}.

\begin{table*}[t]
\centering
\footnotesize
\setlength{\tabcolsep}{3.5pt}
\renewcommand{\arraystretch}{0.99}
\setlength{\aboverulesep}{0pt}
\setlength{\belowrulesep}{0pt}
\begin{tabular}{p{0.35\textwidth}p{0.25\textwidth}>{\centering\arraybackslash}p{0.09\textwidth}>{\centering\arraybackslash}p{0.09\textwidth}}
\toprule
Method & Variant & Medium & Hard \\
\midrule
Full Model & -- & \best{88.0} & \best{81.7} \\
w/o Skill-Specific AMP & -- & 79.5 & 79.3 \\
w/o Skill-Specific Critic & -- & 82.5 & 65.0 \\
\midrule
\multirow{5}{*}{w/o Full Temporal Encoder}
  & Raw Stacked History & 49.0 & 20.0 \\
  & MLP History Encoder & 61.9 & 47.6 \\
  & w/o Ball-Vel Estimation & 83.6 & 44.4 \\
  & w/o Estimation & 79.7 & 15.8 \\
  & Vanilla History Encoder & 78.0 & 32.6 \\
\midrule
w/o Command-Duration Curriculum & -- & 85.5 & 30.4 \\
\bottomrule
\end{tabular}
\caption{\textbf{Ablation Study in Simulation.} Results are overall success rates. The top-level settings ablate the major components; the variants under `w/o Full Temporal Encoder' further isolate the effects of each part.}
\label{tab:ablation_seq}
\end{table*}

\textbf{Results and Analysis.}
Table~\ref{tab:sim_results} evaluates command-driven skill composition in simulation. On the Medium \textsc{Dribble}$\rightarrow$\textsc{Shoot} task, \methodname{} achieves \textbf{88.0\%} overall success, outperforming the best baseline, MoE-Encoder AMP, by 20.8 percentage points. On the Hard task, involving trapping, three-waypoint dribbling, and final shooting, \methodname{} maintains \textbf{81.7\%} overall success, compared with 45.6\% for the best baseline, MoE-Encoder AMP.

\begin{figure*}[t]
    \centering
    \begin{minipage}[t]{0.38\textwidth}
        \vspace{0pt}
        \centering
        \includegraphics[width=0.95\linewidth]{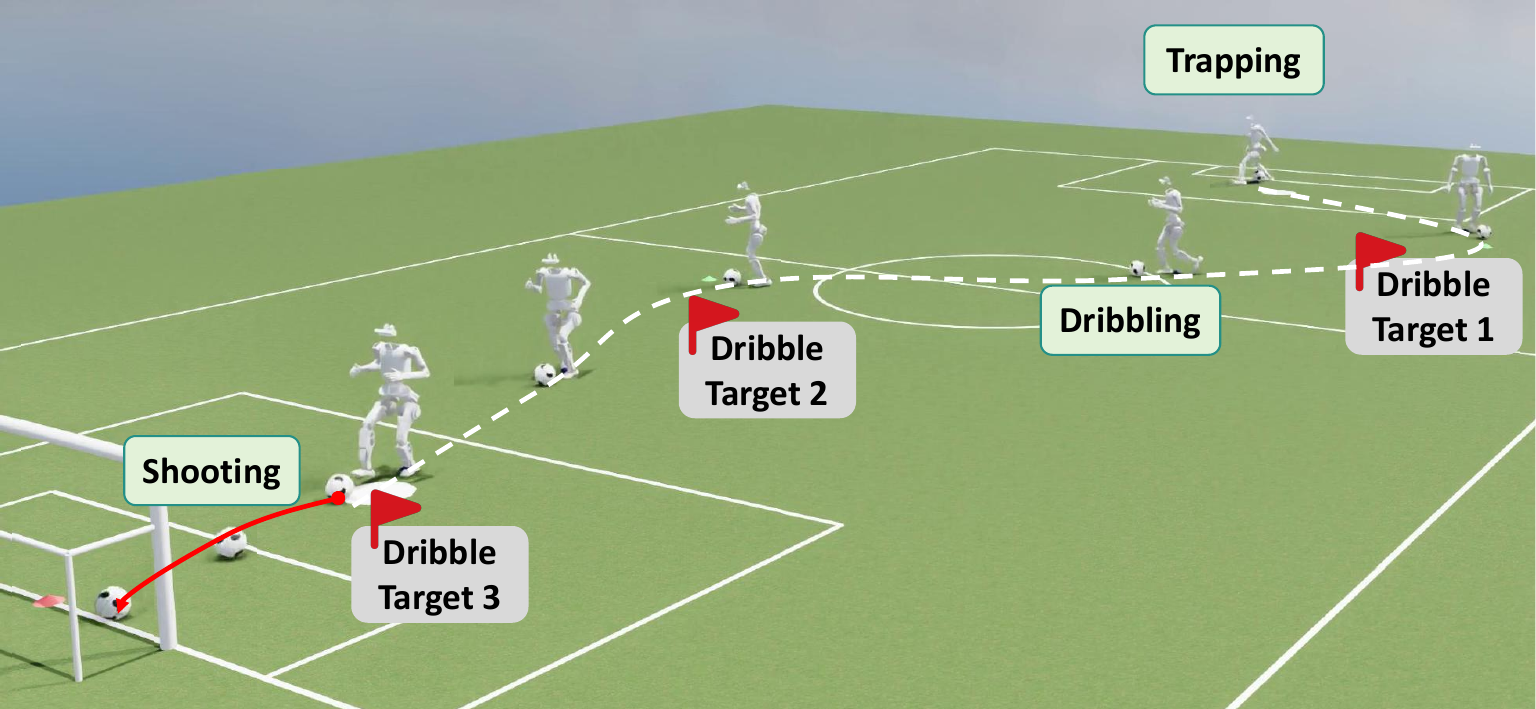}
        \captionsetup{type=figure,font=footnotesize}
        \captionof{figure}{\textbf{Simulation Rollout.}}
        \label{fig:sim_rollout}
    \end{minipage}
    \hfill
    \begin{minipage}[t]{0.58\textwidth}
        \vspace{5pt}
        \centering
        \setlength{\tabcolsep}{4pt}
        \renewcommand{\arraystretch}{1.1}
        \resizebox{\linewidth}{!}{%
            \begin{tabular}{lcccc}
                \toprule
                Method & Dribble & Shoot & Two-step Dribble & Dribble + Shoot \\
                \midrule
                AMP~\cite{peng2021amp} & 1/10 & 1/10 & 0/10 & 0/10 \\
                \rowcolor{gray!25}
                Ours & 8/10 & 8/10 & 7/10 & 6/10 \\
                \bottomrule
            \end{tabular}%
        }
        \captionsetup{type=table,font=footnotesize}
        \captionof{table}{\textbf{Real-World Performance.} Success rates over 10 hardware trials per task under the MoCap system.}
        \label{tab:real_world}
    \end{minipage}
\vspace{-3mm}
\end{figure*}

The substage results further reveal the baselines' failure modes. AMP completes some individual stages but suffers from error accumulation over longer sequences, especially during multi-waypoint dribbling and final shooting. Conditional AMP and MoE-based AMP show unstable composition despite occasional high substage success. MoE-Encoder AMP improves overall completion over these baselines, but its performance still degrades substantially on the Hard task. These results suggest that latent conditioning or expert routing alone does not reliably preserve reusable skills across transitions. In contrast, \methodname{} maintains consistently high substage success and substantially higher overall completion on both composition tasks. 
The Hard-task rollout in Fig.~\ref{fig:sim_rollout} further illustrates how our policy composes trapping, multi-waypoint dribbling, and shooting into a complete long-horizon behavior.

\subsection{Ablation Study}
Table~\ref{tab:ablation_seq} evaluates the major framework components and training strategy, and further decomposes the full temporal-encoder design through five variants: \textbf{(a)} \textit{Raw Stacked History} directly feeds observation history to the actor, without an encoder; \textbf{(b)} \textit{MLP History Encoder} replaces the Transformer with an MLP while retaining the auxiliary objectives; \textbf{(c)} \textit{w/o Ball-Vel Estimation} removes only the ball-velocity estimation objective; \textbf{(d)} \textit{w/o Estimation} removes all velocity estimation objectives; and \textbf{(e)} \textit{Vanilla History Encoder} uses a history MLP without auxiliary objectives. 
Full definitions and controlled settings are provided in Appendix~\ref{app:ablation_detail}.

Removing the skill-specific adversarial motion priors reduces success on both medium and hard tasks, indicating that separated style regularization helps preserve distinct skill behaviors during composition. Removing the skill-specific critics causes a larger drop on the hard task, suggesting that skill-dependent value estimation becomes more important as the task horizon and number of transitions increase.

\textit{Raw Stacked History} drops to 49.0\% on Medium and 20.0\% on Hard despite receiving the same historical inputs, showing that access to history alone is insufficient. \textit{MLP History Encoder} recovers part of the performance but remains below the Full Model, particularly on Hard, indicating the benefit of structured temporal modeling. Removing either all velocity estimation or only ball-velocity estimation also substantially degrades Hard-task performance. Together with the result of \textit{Vanilla History Encoder}, these findings show that the gain arises from the combination of temporal architecture and interaction-relevant auxiliary supervision.

Removing command-duration curriculum causes only a modest decrease on Medium (88.0\% to 85.5\%) but a substantial drop on Hard (81.7\% to 30.4\%), indicating that the curriculum is especially important for long-horizon skill composition.

\begin{figure*}[t]
    \centering
    \includegraphics[width=0.99\textwidth]{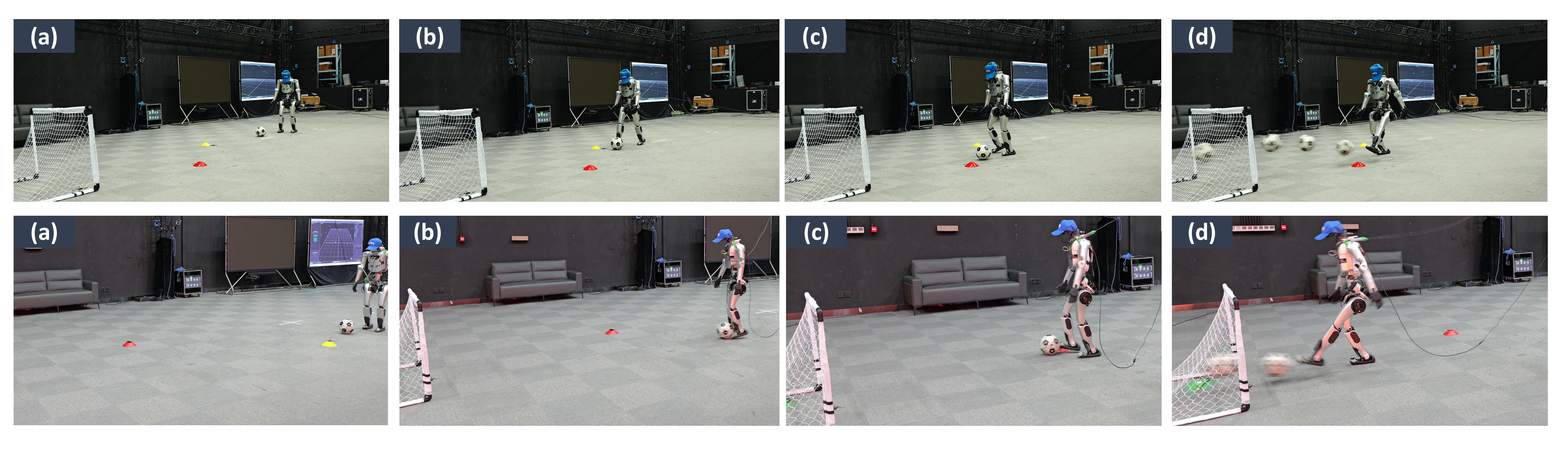}
    \caption{\textbf{Real-World Rollouts} on the Noetix E1 humanoid robot under the MoCap system.}
    \label{fig:real_rollouts}
\vspace{-5mm}
\end{figure*}

\subsection{Real-World Experiments}

\textbf{Evaluation Setting.}
We evaluate real-world performance under the MoCap deployment setup
described in Sec.~\ref{app:mocap}. The evaluation contains four tasks of varying difficulty: dribbling to a target region, shooting to the goal,
two-step dribbling with a direction change, and dribble-then-shoot. 
The first two tasks evaluate atomic skill execution, while the latter two
evaluate long-horizon skill composition and command-driven transitions.
For the composite hardware tasks, skill commands are manually triggered
when the corresponding subtask is completed.

\textbf{Metrics and Baseline.} We report success over 10 hardware trials per task. A trial is successful if the robot completes the task objective without falling and the
ball remains within the valid field boundary. For composite tasks, all subtasks
must be completed in the specified order. We compare \methodname{} with the standard AMP baseline using the same deployment interface and sensing pipeline.

\textbf{Results.}
Table~\ref{tab:real_world} shows that \methodname{} transfers reliably to the real robot and substantially outperforms AMP across all tasks. On atomic skills, our policy achieves \textbf{80\%} success on both dribbling and shooting, whereas AMP succeeds in only 1 out of 10 trials for each task. The performance gap becomes larger on composite tasks: \methodname{} achieves \textbf{70\%} and \textbf{60\%} success, respectively, while AMP fails in all trials. These results indicate that separating skill-specific training signals improves not only individual skill execution but also real-world robustness during command-driven skill transitions. Qualitative rollouts in Fig.~\ref{fig:real_rollouts} further illustrate that the policy can execute robust soccer skill behaviors on hardware.

\begin{figure}[t]
    \centering
    \includegraphics[width=\textwidth]{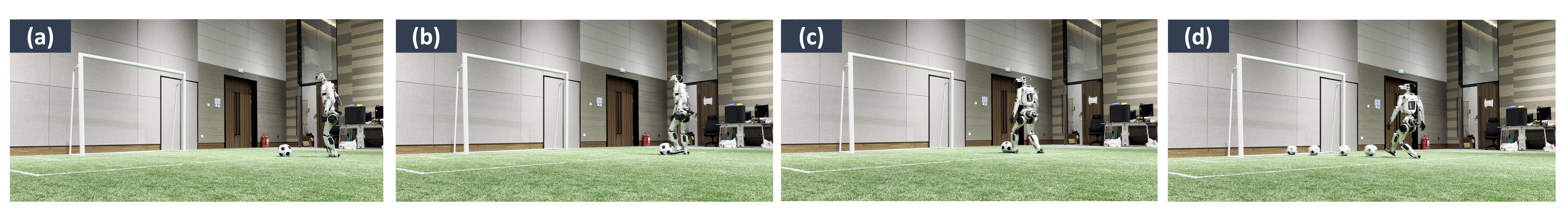}
    \caption{\textbf{Onboard vision deployment} of our policy on the dribble-then-shoot composite task. }
    \label{fig:yolo_rollout}
\end{figure}

\begin{figure}[t]
    \centering
    \includegraphics[width=0.98\textwidth]{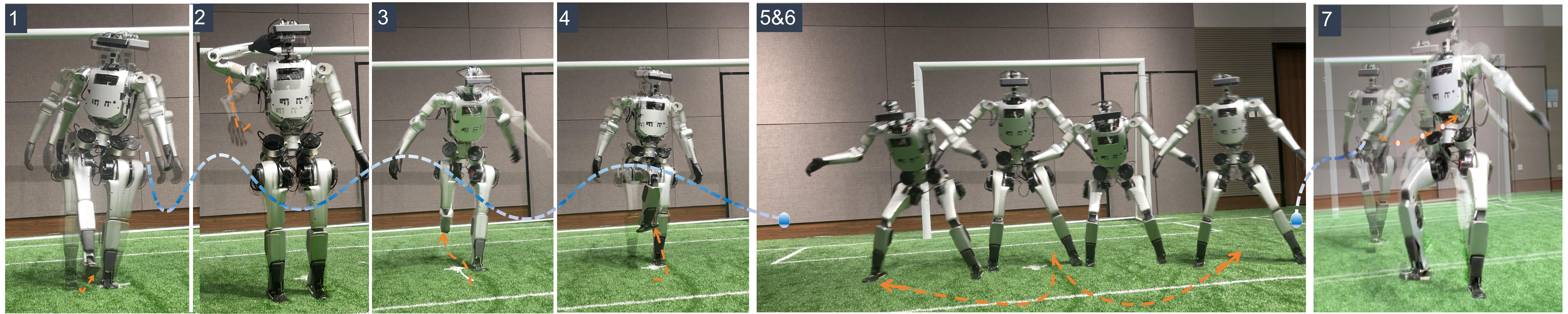}
    \caption{
    \textbf{Smooth Skill Transitions within One Policy.}
    A single command-conditioned policy executes and transitions among diverse whole-body skills, including (1) \textit{kick}, (2) \textit{salute}, (3) \textit{back kick}, (4) \textit{single-leg stand}, (5-6) \textit{left/right defending motions}, and (7) \textit{run-to-shoot}.
    }
    \label{fig:no_ball}
    \vspace{-5mm}
\end{figure}

\subsection{Onboard Vision Deployment}

As an additional feasibility study beyond the MoCap evaluation, we deploy
\methodname{} with onboard perception using a head-mounted ZED2i camera with
visual-inertial odometry and YOLOv8-based ball detection. The estimated robot
and ball states are converted into the same robot-centric policy interface used
in simulation and MoCap deployment. To account for perception noise, we train
with degraded observations, including lower-rate updates and observation delays.
Although onboard vision is noisier than MoCap, the policy can still execute ball-interaction behaviors, as shown in Fig.~\ref{fig:yolo_rollout}, suggesting improved sim-to-real robustness under practical sensing conditions. More qualitative results and deployment details are provided in Appendix~\ref{app:yolo}. 

\subsection{Skill Diversity Beyond Soccer Interaction}

Beyond ball-interaction tasks, we further evaluate \methodname{} on multi-skill stylized motion imitation. We remove external object observations and object-interaction rewards, so the policy is driven only by proprioception and skill commands while retaining the same training framework. As shown in Fig.~\ref{fig:no_ball}, a single policy learns diverse whole-body skills and transitions smoothly among them, including kicking, saluting, back kicking, single-leg standing, defending motions, and run-to-shoot. This result shows that the \methodname{} design is not tied to ball-interactive tasks and can support broader multi-skill humanoid motion learning.

\section{Conclusion}
\label{sec:conclusion}

We presented \textbf{\methodname{}}, a unified reinforcement learning framework for \textbf{multi-skill humanoid soccer}. \methodname{} uses a single command-conditioned policy for deployment while using \textbf{skill-specific adversarial motion priors}, \textbf{skill-specific critics}, and an \textbf{object-aware temporal encoder} to address style interference, skill-dependent value estimation, and partial observability. Experiments in simulation and on a real Noetix E1 humanoid demonstrate improved atomic skill execution, robust sim-to-real deployment, and \textbf{long-horizon multi-skill composition}. These results suggest that combining a unified deployable policy with skill-specialized training signals is a promising direction for multi-skill object-interactive control.

\section{Limitations}

\methodname{} focuses on low-level multi-skill execution and command-conditioned transitions under externally provided high-level skill commands. Therefore, it does not address strategic decision-making, task planning, or autonomous skill sequencing in open-ended soccer scenarios. Our experiments also focus on single-robot tasks; extending the framework to multi-agent soccer with teammates and opponents would introduce additional challenges such as physical disturbances, coordination, and strategic skill selection. Future work may integrate \methodname{} with hierarchical planning or multi-agent coordination to support broader soccer scenarios.

\clearpage
\acknowledgments{We thank Beijing Virtual Point Tech Co., Ltd. for providing motion-capture support for the real-world experiments.}


\bibliography{example}  

\clearpage
\appendix

\section{Deployment Details}
\label{app:deploy}
Our real-world experiments are conducted on the 25-DoF Noetix E1 humanoid robot. The learned policy runs on the onboard computer and outputs target joint positions at 50 Hz, which are tracked by the low-level motor controller. For deployment safety, policy actions are clipped before being sent to the low-level controller. We implement two real-world deployment backends: a motion-capture backend for controlled evaluation and an onboard-vision backend for infrastructure-free execution.

\textbf{Unified deployment interface.}
Both deployment backends use the same robot-centric policy interface. Raw measurements from MoCap, onboard vision, or simulation are converted into a common observation format containing ball position, target position, proprioception, previous action, and skill command. The observation is appended to a short history buffer and fed to the exported ONNX policy. This design separates the learned policy from the sensing backend and reduces deployment-specific changes.

\subsection{Motion-Capture Deployment}
\label{app:mocap}

For MoCap deployment, rigid-body poses from the Motive system are received through VRPN and exposed to the controller as ROS topics. The MoCap system provides global measurements of the robot, ball, and goal. We attach a rigid 3D-printed marker mount to the robot torso and calibrate its fixed transform to the robot \(\texttt{base\_link}\), allowing the measured rigid-body pose to be converted into the robot base frame. Reflective markers are also attached to the ball and goal to estimate their global positions.

The controller transforms all global object measurements into robot-centric observations. Let $\mathbf{p}^{w}_{r}$ and $\mathbf{R}^{w}_{r}$ denote the robot position and orientation in the world frame, and let $\mathbf{p}^{w}_{o}$ denote the world-frame position of an external object such as the ball or target. 
\begin{equation}
    \mathbf{p}^{r}_{o} = (\mathbf{R}^{w}_{r})^\top \left(\mathbf{p}^{w}_{o} - \mathbf{p}^{w}_{r}\right),
\end{equation}
Only the xy-plane components are used for ball and target observations.

The MoCap system runs at 120 Hz, while the policy runs at 50 Hz. Although this setup provides repeatable closed-loop evaluation, it is not equivalent to a perfect simulator state: relative object observations can be affected by tracking jitter, temporary occlusion, calibration error, and latency, especially for fast ball motion. We therefore apply exponential low-pass filtering to relative object observations and treat stale rigid-body measurements as invalid to avoid using outdated data. The marker mount is positioned to reduce occlusion during dynamic motions and improve tracking reliability.

Fig.~\ref{fig:real_rollouts} illustrates representative deployment rollouts under the MoCap setup.
Utilizing high-precision, real-time object state estimations from the tracking infrastructure, the robot robustly switches between distinct skills. This continuous feedback loop facilitates seamless closed-loop object interactions, ensuring high-fidelity coordination and successful completion of the final tactical task.

\subsection{Onboard Vision Deployment}
\label{app:yolo}

For onboard-vision deployment, we use a head-mounted ZED2i camera. Robot odometry is estimated from the camera's visual-inertial odometry module, while the soccer ball is detected using a YOLOv8-based detector. The detector runs at approximately 30 Hz with GPU acceleration. The 2D detection is converted into a robot-centric 3D ball observation using depth measurements and camera calibration. Since the target location is fixed in the field, its relative position is obtained from the estimated odometry. The resulting ball and target observations are converted into the same robot-centric policy interface used in MoCap deployment.

To improve robustness to onboard perception, we train the policy with a perception-degraded observation model rather than directly using perfect object states. During training, the object observation is updated at a lower rate, perturbed with noise and random delay, and temporarily frozen when the ball is outside the simulated camera field of view. We also simulate basic geometric visibility constraints, including camera range and field-of-view limits, to approximate the failure modes of the physical detector. These perturbations encourage the policy to handle delayed, noisy, and intermittent object observations during deployment.


Fig.~\ref{fig:yolo_rollout} shows representative onboard-vision deployment rollouts. The robot can localize the ball from the head-mounted camera, approach it, and execute closed-loop ball-interaction behaviors without relying on an external MoCap system. Although this setup is noisier and less reliable than MoCap-based sensing, the same policy interface can accommodate both sensing backends, demonstrating the robustness of our deployment design under practical perception conditions.

\section{Simulation Experiment}
\subsection{Evaluation Protocol}
\label{app:task_detail}
We evaluate the proposed method in NVIDIA Isaac Lab~\cite{mittal2025isaaclab} using a soccer-field environment. 
All policies are evaluated with a single environment, and each episode is executed sequentially. 
For each task setting, we report the success rate over independent evaluation episodes and the average completion time over successful episodes. 
During evaluation, all domain randomization and observation corruption are disabled to ensure reproducibility.

The evaluation field is a $0.25\times$ scaled standard soccer field with a length of $25.0\,\mathrm{m}$ and a width of $15.0\,\mathrm{m}$. 
The goal width and height are $2.745\,\mathrm{m}$ and $1.22\,\mathrm{m}$, respectively. 
The ball radius is $0.11\,\mathrm{m}$ and its mass is fixed to $0.45\,\mathrm{kg}$ during evaluation.

We consider three atomic-skill tasks and two composite tasks. Atomic-skill tasks are designed to isolate a single soccer primitive at a time. In these tasks, the high-level skill command is fixed throughout the episode, and the policy is only required to complete one behavior. This setting provides a controlled evaluation of the learned motion primitives before they are composed into longer and more complex soccer tasks.

\textbf{Trap.}
The trap task evaluates the robot's ability to stop an incoming ball. 
The robot is initialized at the origin and faces the positive $x$ direction. 
The ball is initialized $1.0$--$2.0\,\mathrm{m}$ away from the robot with a random bearing within $\pm 0.785\,\mathrm{rad}$ and an initial speed of $0.5$--$1.5\,\mathrm{m/s}$ toward the robot. 
The skill command is fixed to \textsc{Trap}. 
The task is considered successful when the ball's planar velocity becomes lower than $0.3\,\mathrm{m/s}$ within $20\,\mathrm{s}$.

\textbf{Dribble.}
The dribble task evaluates the robot's ability to move the ball to a randomly sampled target position. 
The robot is initialized at the origin, and the ball is initialized $0.5$--$1.0\,\mathrm{m}$ away with an angular offset within $\pm \pi/3$. 
The target is sampled with $x \in [4.5, 5.5]\,\mathrm{m}$ and $y \in [-2.0, 2.0]\,\mathrm{m}$. 
The skill command is fixed to \textsc{Dribble}. 
The task is successful if the ball reaches within $0.5\,\mathrm{m}$ of the target within $20\,\mathrm{s}$.

\textbf{Shoot.}
The shoot task evaluates the robot's ability to kick a stationary ball into the goal. 
The robot is initialized $4$--$6\,\mathrm{m}$ in front of the goal with a heading perturbation within $\pm 0.3\,\mathrm{rad}$. 
The ball is initialized $0.5$--$1.2\,\mathrm{m}$ away from the robot with an angular offset within $\pm 0.3\,\mathrm{rad}$. 
The skill command is fixed to \textsc{Shoot}. 
An episode is successful if the ball crosses the goal line, and it fails after $15\,\mathrm{s}$ if the goal is not reached.

We further evaluate the policy on composed tasks that explicitly require multi-stage skill execution.

\textbf{Medium.}
The medium task evaluates a composed behavior consisting of dribbling followed by shooting. 
The robot starts near the midfield with the ball initialized $0.6$--$1.2\,\mathrm{m}$ away. 
The initial skill command is \textsc{Dribble}. 
After the ball reaches the waypoint at the penalty point, the command is automatically switched to \textsc{Shoot}. 
The episode is successful if the ball enters the goal within $30\,\mathrm{s}$.

\textbf{Hard.}
The hard task evaluates the full composition of trapping, dribbling, and shooting. 
The task starts with the \textsc{Trap} skill, where the ball moves toward the robot with an initial speed of $0.5$--$1.0\,\mathrm{m/s}$. 
Once the ball speed falls below $0.3\,\mathrm{m/s}$, the skill command switches to \textsc{Dribble}. 
The robot then needs to guide the ball through three waypoints:
$(-6.25, 1.145)$, $(0, -1.145)$, and $(9.75, 0)$ along an S-shaped route. 
After all waypoints are reached, the command switches to \textsc{Shoot}. 
The task is successful if the ball enters the goal within $40\,\mathrm{s}$.

\subsection{Baseline Settings}
\label{app:baseline}
We compare our method with five baselines: Vanilla PPO, AMP, Conditional AMP, MoE-based AMP, and MoE-Encoder AMP. Conditional AMP, MoE-based AMP, and MoE-Encoder AMP are self-implemented in our framework. 
All methods use the same robot model, simulation settings, observation space, task reward design, reference motion dataset, and evaluation protocol.

\textbf{Vanilla PPO.}~\cite{schulman2017proximal}
Vanilla PPO is trained only with task rewards and does not use adversarial motion priors. 
The actor and critic are both multilayer perceptrons with hidden dimensions 
$[2048, 1024, 1024, 512, 512]$ and ELU activations. 

\textbf{AMP.}~\cite{peng2021amp}
AMP uses a standard adversarial motion prior with a single discriminator shared by all skills. 
The policy and value networks follow the same architecture as Vanilla PPO. 
The discriminator is an MLP with hidden dimensions $[1024, 512]$. 
The AMP reward coefficient is set to $1.0$.

\textbf{Conditional AMP.}~\cite{huang2025learning}
Conditional AMP extends AMP with latent-based skill conditioning. It maps the skill command $g$ to a target latent $z=E(g)$. In addition to the shared AMP discriminator, a skill discriminator with hidden dimensions $[512, 256]$ predicts $\hat z=f_\psi(\tau)$ from motion trajectory $\tau$ and provides the cosine-similarity reward $r^{\mathrm{skill}}=\cos(\hat z,z)$. The total reward is $r=r^{\mathrm{task}}+\lambda^{\mathrm{style}}r^{\mathrm{style}}+\lambda^{\mathrm{skill}}r^{\mathrm{skill}}$. The AMP style reward coefficient is $1.0$, and the skill reward coefficient is $0.5$.

\textbf{MoE-based AMP.}~\cite{huang2025moe}
MoE-based AMP combines a mixture-of-experts policy with AMP. 
It uses three policy experts, corresponding to the number of skills, and a gating network with a hidden dimension of $128$. 
Top-1 routing is used, meaning that only one expert is activated at each step. 
Each expert has its own actor and critic networks. 

\textbf{MoE-Encoder AMP.}~\cite{wu2026toward}
MoE-Encoder AMP augments AMP with a mixture-of-experts history encoder using soft-gated dense routing. The encoder contains three experts and a soft gating network, and maps a five-step observation history to a 32-dimensional $\ell_2$-normalized latent representation. The resulting latent is concatenated with the current observation and provided to the actor. We adopt the load-balancing objective from the original paper~\cite{wu2026toward} for expert utilization, but do not use CTS training or RoboGauge evaluation protocol.

\paragraph{Shared hyperparameters.}
All baseline methods are trained under the same PPO hyperparameter configuration, except for method-specific network structures and adversarial modules. 
The shared training hyperparameters are summarized in Table~\ref{tab:shared_hyperparameters}.
\begin{table}[h]
    \centering
    \caption{\textbf{Shared training hyperparameters.} These hyperparameters are kept the same for all baseline methods and our method unless otherwise specified.}
    \vspace{2mm}
    \label{tab:shared_hyperparameters}
    \begin{tabular}{lc}
        \toprule
        Hyperparameter & Value \\
        \midrule
        Learning epochs & 5 \\
        Mini-batches & 4 \\
        Learning rate & $1.0 \times 10^{-3}$, adaptive schedule \\
        Discount factor $\gamma$ & 0.99 \\
        GAE parameter $\lambda$ & 0.95 \\
        Entropy coefficient & 0.01 \\
        Target KL divergence & 0.01 \\
        Gradient clipping & 1.0 \\
        PPO clipping coefficient & 0.2 \\
        Value loss coefficient & 1.0 \\
        Action clipping & 18.0 \\
        Initial noise standard deviation & 1.0, scalar \\
        \bottomrule
    \end{tabular}
\end{table}

\subsection{Details of Ablation Experiments}
\label{app:ablation_detail}
We conduct simulation-based ablation studies to quantify the contribution of each core component in our framework.

The framework-level variants are defined as follows: w/o Skill-Specific AMP replaces the skill-specific discriminators with a single discriminator, w/o Skill-Specific Critic uses a single shared critic head instead of skill-specific value heads, and w/o Command-Duration Curriculum samples the final command-duration range from the beginning of training instead of progressively shortening skill segments. 

The five variants under w/o Full Temporal Encoder further isolate the source of the complete encoder design's gains.
\textbf{(a)} \textit{Raw Stacked History} preserves the full five-frame observation history but flattens it and feeds it directly to the actor, without any encoder or encoder-side auxiliary objective.
\textbf{(b)} \textit{MLP History Encoder} replaces the Transformer with an MLP while retaining the same auxiliary objectives.
\textbf{(c)} \textit{w/o Ball-Vel Estimation} removes the ball-velocity prediction target while retaining the remaining encoder objectives.
\textbf{(d)} \textit{w/o Estimation} removes all velocity-estimation objectives while retaining the other parts of the temporal encoder, whereas \textbf{(e)} \textit{Vanilla History Encoder} uses a history MLP without auxiliary objectives.
These comparisons distinguish the effects of an explicit history representation, temporal architecture, and interaction-relevant supervision.

\subsection{Additional Results}
\label{app:sim_results}

In addition to the composite task results reported in the main text, we further evaluate all methods on isolated atomic skill tasks. These results serve to verify whether the learned policy can reliably execute each primitive skill. Each atomic-skill result is evaluated over 100 trials per method.

Table~\ref{tab:single_skill_eval} shows that our method achieves the best overall performance across the three atomic skills. Vanilla PPO performs well on trapping but fails on shooting and dribbling, indicating that task rewards alone are insufficient for highly dynamic, contact-rich ball skills. Although Standard AMP improves these skills with motion priors, its shared discriminator still underperforms our method. Conditional AMP and MoE-based AMP also fail to consistently learn all skills, suggesting that skill conditioning or expert routing cannot effectively resolve interference among heterogeneous skills. In contrast, our method achieves \textbf{95.0\%} success on shooting and \textbf{100.0\%} success on both trapping and dribbling, demonstrating more balanced and robust primitive-skill learning.

\begin{table}[h]
\centering
\small
\caption{\textbf{Atomic-Skill tasks for baseline experiments.}}
\label{tab:single_skill_eval}
\setlength{\tabcolsep}{7pt}
\begin{threeparttable}
\begin{tabular}{l cc cc c}
\toprule
\multirow{2}{*}{Method} & Shoot & \multicolumn{2}{c}{Trap} & \multicolumn{2}{c}{Dribble}  \\
\cmidrule(lr){2-2} \cmidrule(lr){3-4} \cmidrule(lr){5-6}
 & Succ.(\%) & Succ.(\%) & Time(s) & Succ.(\%) & Time(s)  \\
\midrule
Vanilla PPO & 9.0 & 97.0 & \best{3.90} & 5.0 & 7.26  \\
AMP & 87.0 & 99.0 & 5.22 & 94.0 & 6.27  \\
Conditional AMP & 1.0 & 100.0 & 5.84 & 8.0 & 6.70  \\
MoE-based AMP & 1.0 & 99.0 & 5.77 & 36.0 & 5.90  \\
\rowcolor{gray!25}
\textbf{Ours} & \best{95.0} & \best{100.0} & 5.31 & \best{100.0} & \best{5.62}  \\
\bottomrule
\end{tabular}
\end{threeparttable}
\end{table}

Table~\ref{tab:atomic_skill} further reports ablation experiments on atomic-skill tasks. 
Removing skill-specific AMP mainly degrades shooting performance, showing that skill-specific adversarial supervision is particularly important for dynamic contact behaviors. 
Removing the skill-specific critics also reduces shooting success, indicating that skill-specific value estimation helps stabilize optimization across heterogeneous skills. 
Raw Stacked History causes a substantial drop in shooting and dribbling performance despite retaining the five-frame inputs, suggesting that explicitly encoding temporal context is important for stable ball control.
Overall, these supplementary results confirm that the components of our method not only improve the complex multi-stage tasks in the main text, but also strengthen the underlying atomic skills required for successful skill composition.

\begin{table}[h]
\centering
\caption{\textbf{Atomic-Skill tasks for ablation experiments.}}
\label{tab:atomic_skill}
\begin{tabular}{lccccc}
\toprule
\multirow{2}{*}{Method}
& Shoot
& \multicolumn{2}{c}{Trap}
& \multicolumn{2}{c}{Dribble} \\
\cmidrule(lr){2-2}
\cmidrule(lr){3-4}
\cmidrule(lr){5-6}
& Succ.(\%)
& Succ.(\%) & Time(s)
& Succ.(\%) & Time(s) \\
\midrule
w/o Skill-Specific AMP & 87.0 & 100.0 & 2.77 & 100.0 & 5.60 \\
w/o Skill-Specific Critic & 85.0 & 100.0 & 5.32 & 100.0 & 6.66 \\
Raw Stacked History & 16.0 & 98.0 & 5.77 & 91.0 & 10.80 \\
\rowcolor{gray!25}
\textbf{Ours} & \best{95.0} & \best{100} & 5.31 & \best{100.0} & 5.62\\
\bottomrule
\end{tabular}
\end{table}

\paragraph{Stricter trapping criterion.}
To further evaluate trapping under a more challenging criterion, we measure success from the first robot--ball contact, defined by a planar distance of at most $0.4\,\mathrm{m}$. A trial is successful only if, within $3\,\mathrm{s}$ after contact, the ball maintains a speed below $0.25\,\mathrm{m/s}$ and remains within $0.5\,\mathrm{m}$ of the robot for at least $0.5\,\mathrm{s}$. Over 100 trials per method, \methodname{} achieves a success rate of \textbf{63.0\%}, compared with 31.0\% for AMP, 24.0\% for Conditional AMP, and 50.0\% for MoE-based AMP, retaining the best performance among all compared methods under this stricter evaluation.

\section{Implementation Details}
\label{app:impl_detail}

\subsection{Observation Details}
\label{app:obs_detail}
We use three observation groups during training: actor observations, critic observations, and discriminator observations. The actor observations are available during both training and deployment, while the critic and discriminator observations are used only during training.

\paragraph{Actor observation.}
The actor is fed with a historical observation window of length \(K=5\). Each timestep frame integrates proprioceptive states, external object observations, and the skill command, formulated as:
\begin{align}
\mathbf{o}_t &=
    \left[
    \mathbf{p}_{t},
    \mathbf{c}_{t},
    \mathbf{o}^{\mathrm{ext}}_{t}
    \right], \\
    \mathbf{p}_{t} &= \left[\boldsymbol{\omega}^{\mathrm{base}}_t,
\mathbf{g}^{\mathrm{body}}_t,
\mathbf{q}_t,
\dot{\mathbf{q}}_t,
\mathbf{a}_{t-1}\right], \\
\mathbf{o}^{\mathrm{ext}}_{t} &= \left[\mathbf{p}^{\mathrm{ball}}_{t,xy}, \mathbf{p}^{\mathrm{target}}_{t,xy}\right].
\end{align}

Here, \(\boldsymbol{\omega}^{\mathrm{base}}_t\in\mathbb{R}^{3}\) denotes the base angular velocity, \(\mathbf{g}^{\mathrm{body}}_t\in\mathbb{R}^{3}\) represents the gravity vector projected onto the body frame, \(\mathbf{q}_t,\dot{\mathbf{q}}_t\in\mathbb{R}^{25}\) correspond to joint positions and velocities respectively, \(\mathbf{a}_{t-1}\in\mathbb{R}^{25}\) is the prior-step action, \(\mathbf{p}^{\mathrm{ball}}_{t,xy}\in\mathbb{R}^{2}\) defines the horizontal-plane ball position relative to the robot, \(\mathbf{p}^{\mathrm{target}}_{t,xy}\in\mathbb{R}^{2}\) specifies the horizontal-plane target position relative to the robot, and \(\mathbf{c}_t\in\mathbb{R}^{N_{\mathrm{skill}}}\) stands for the skill command.

Accordingly, the dimensionality of the per-frame actor observation is calculated as:$
d_{\pi}=3+3+25+25+25+2+2+N_{\mathrm{skill}}
=85+N_{\mathrm{skill}}$.
For our core soccer experiments with \(N_{\mathrm{skill}}=3\), the per-frame dimension reduces to \(d_{\pi}=88\). Combined with the history window \(K=5\), the actor input features a shape of \(5\times 88\), amounting to 440 scalar values when flattened. During training, observation noise is injected into all components of the actor observation except the skill command.

\paragraph{Critic observation.}
Following the standard asymmetric actor-critic paradigm, the critic adopts a privileged observation design that differs from the actor. The critic takes the uncorrupted actor observation sequence as the base input, and further incorporates additional privileged state information exclusive to the training phase:
\begin{equation}
\mathbf{o}^{V}_t =
\left[
\mathbf{o}_t,
\mathbf{v}^{\mathrm{base}}_t,
\mathbf{v}^{\mathrm{ball}}_t,
\mathbf{f}^{\mathrm{feet}}_t
\right].
\end{equation}

Here, \(\mathbf{v}^{\mathrm{base}}_t\in\mathbb{R}^{3}\) denotes the base linear velocity of the robot, \(\mathbf{v}^{\mathrm{ball}}_t\in\mathbb{R}^{3}\) represents the ball velocity expressed in the robot-centric frame, and \(\mathbf{f}^{\mathrm{feet}}_t\in\mathbb{R}^{2}\) refers to the binary contact state of the robot’s two feet. The dimensionality of the per-frame critic observation is thus formulated as: $d_V=d_{\pi}+3+3+2=93+N_{\mathrm{skill}}$. In our soccer experimental setup with \(N_{\mathrm{skill}}=3\), the critic observation dimension yields \(d_V=96\). Consistent with the asymmetric actor-critic framework, these privileged observational features are solely utilized for value function estimation during model training, and are completely discarded in the deployed policy, incurring no extra inference overhead for the actor.

\paragraph{Discriminator observation.}
The AMP discriminator adopts a set of motion-centric observations distinct from the actor and critic modalities, which focus on capturing full-body motion characteristics for adversarial training. The discriminator observation is formulated as:
\[
\mathbf{o}^{D}_t =
\left[
\mathbf{q}_t,
\mathbf{x}^{\mathrm{body}}_t,
\mathbf{v}^{\mathrm{base}}_t,
\boldsymbol{\omega}^{\mathrm{base}}_t,
\dot{\mathbf{q}}_t,
\mathbf{f}^{\mathrm{feet}}_t
\right].
\]

Here, \(\mathbf{q}_t\in\mathbb{R}^{25}\) and \(\dot{\mathbf{q}}_t\in\mathbb{R}^{25}\) represent the joint positions and joint velocities, respectively. \(\mathbf{x}^{\mathrm{body}}_t\in\mathbb{R}^{3N_{\mathrm{body}}}\) encodes the positional information of all robot rigid bodies, where \(N_{\mathrm{body}}\) denotes the total number of robot bodies. For the 25-DoF Noetix E1 robot adopted in our experiments, \(N_{\mathrm{body}} = 26\). In addition, \(\mathbf{v}^{\mathrm{base}}_t\in\mathbb{R}^{3}\) is the base linear velocity, \(\boldsymbol{\omega}^{\mathrm{base}}_t\in\mathbb{R}^{3}\) is the base angular velocity, and \(\mathbf{f}^{\mathrm{feet}}_t\in\mathbb{R}^{2}\) corresponds to the binary foot-contact states of the robot’s two feet.

The dimensionality of the per-frame discriminator observation is calculated as $
d_D
=
25 + 3 \times 26 + 3 + 3 + 25 + 2
=136.$

The discriminator is trained on sequential multi-step motion segments constructed from the above motion-centric observations. Discriminator observations are leveraged for adversarial motion prior training during the training phase, without being involved in policy deployment.

\subsection{Training Strategy}
\label{app:strategy}

\paragraph{Adaptive Command Sampling.}
We employ an adaptive command sampling strategy to prioritize underperforming skills and alleviate imbalanced learning. The failure rate \(f_i\) of each skill is tracked via exponential moving average (EMA) with a smoothing coefficient of \(0.99\). The sampling probability for skill \(i\) is computed as a weighted mixture between the failure-aware distribution and a uniform prior:
\begin{align}
p_i &= (1-\alpha) \cdot \frac{f_i}{\sum_j f_j} + \alpha \cdot \frac{1}{N_{\text{skill}}},
\end{align}
where the mixing weight \(\alpha\) is set to \(0.5\) to balance adaptive focusing and uniform exploration. An initial warm-up phase of \(5000\) iterations uses uniform random sampling for all skills. After the warm-up period, the adaptive sampling strategy is activated to allocate more updates to skills with higher failure rates.

\paragraph{Command Duration Curriculum.}
A command duration curriculum is designed to gradually improve the policy's ability to perform frequent and smooth skill transitions. Before iteration \(9000\), the command duration is uniformly sampled from a fixed range \((5.0, 10.0)\) seconds to encourage stable single-skill execution. From iteration \(9000\) to \(12000\), the lower bound of the duration range is linearly annealed from \(5.0\) seconds to \(0.2\) seconds, while the upper bound remains fixed at \(10.0\) seconds. After iteration \(12000\), the duration range is fixed at \((0.2, 10.0)\) seconds for the remainder of training. This curriculum forces the policy to first master stable behaviors and then learn agile skill switching.

\paragraph{Object Domain Randomization.}
To improve the policy's robustness against physical variations during real-world deployment and reduce the sim-to-real gap in ball-interactive tasks, we apply domain randomization~\cite{peng2018sim} to the material and mass properties of the soccer ball at each environment reset. Specifically, we randomize the static friction coefficient within \((0.30, 0.70)\), the dynamic friction coefficient within \((0.10, 0.55)\), and the restitution within \((0.55, 0.95)\). The ball mass is randomized in the range of \((0.40, 0.48)\) kg. All physical properties are sampled independently and clamped to the specified ranges. This domain randomization improves generalization to varying object interaction dynamics without modifying the core learning framework.

\paragraph{Command Blending.}
Command blending is applied only during the 5-frame ($0.1\,\mathrm{s}$) transition window, accounting for approximately $2\%$ of training steps. Therefore, the blended value is used as a local approximation to smooth value changes near skill transitions; it does not assume that the returns of arbitrarily interpolated policies are globally linear.

\end{document}